# A New Algorithm for Finding MAP Assignments to Belief Networks*

Solomon E. Shimony and Eugene Charniak
Computer Science Department
Box 1910, Brown University
Providence, RI 02912
ses@cs.brown.edu and ec@cs.brown.edu

## Abstract

We present a new algorithm for finding maximum a-posteriori (MAP) assignments of values to belief networks. The belief network is compiled into a network consisting only of nodes with boolean (i.e. only 0 or 1) conditional probabilities. The MAP assignment is then found using a best-first search on the resulting network. We argue that, as one would anticipate, the algorithm is exponential for the general case, but only linear in the size of the network for poly trees.

## Introduction

Algorithms for belief networks (Bayesian networks) are the cornerstone of many applications for probabilistic reasoning. Effective algorithms exist for calculating posterior probabilities of nodes given the evidence, even in the case of undirected cycles in the network. Some of these are based on Pearl's message passing algorithms (see [Pearl, 1988]), where some preprocessing is needed, such as clustering or conditioning.

While much has been written about finding posterior probabilities of nodes, not much has been done about finding maximum probability assignments (MAPs) for belief networks[1]. One algorithm to compute MAPs is given by Pearl in [Pearl, 1988]. That algorithm, however is rather complicated, and finding the next best assignments with that algorithm is not as simple as with the algorithm we present.

Cooper, in his PhD thesis (see [Cooper, 1984], or [Neapolitan, 1990]), performs a best-first search for a most probable set of "diseases", or causes, given the evidence. That, however, is not equivalent to calculating a complete MAP, as he assumes mutual independence of all causes (i.e. they all have to be root nodes). Peng and Reggia, in [Peng and Reggia, 1987], have defined a diagnostic problem that uses a 2-level belief network, and designed a best-first algorithm that finds hypotheses in decreasing order of probability. It is not clear, however, how their methods would extend to a general belief network, given that one of their assumptions is that all symptoms have causes (thus root nodes cannot be evidence).

We propose an algorithm that transforms the belief net into a weighted boolean-function DAG, and then performs a best-first search to find the least cost assignment, which induces the MAP assignment on the belief network (it can find any next-best assignments as a natural extension). In the next section we define our transformation, and show that a minimum cost assignment for the cost-based DAG induces a MAP assignment on the belief network. In sections following that, we describe the algorithm and discuss complexity issues. We then present some experimental results of using two variants of the algorithm for limited belief networks, and conclude with a summary of our results and a discussion of future research.

## Belief Nets as Weighted DAGS

In this section, we define weighted boolean function DAGs (WBFDAGs), and show how to represent any given Bayesian net as a WBFDAG. We assume that the Bayesian network uses only discrete random variables. We also assume, without loss of generality, that all nodes take on the same values[2], i.e. values from the domain $\mathcal{D} = \{L_1, L_2, ..., L_m\}$.

A WBFDAG is a DAG where nodes can be assigned values from some domain $\mathcal{D}'$. Nodes have labels, which are functions in $\mathcal{F}$, the set of all functions with domain $\mathcal{D}'^k$ for some $k$, and range $\mathcal{D}'$. Formally, we define such DAGs as follows:

---

*This work has been supported in part by the National Science Foundation under grants IRI-8911122 and Office of Naval Research under grant N00014-88-K-0589. We wish to thank Robert Goldman for many helpful comments.

[1]MAP stands for "Maximum A-posteriori Probability", and we use it to refer to a *complete* assignment, unless specified otherwise.

[2]If this is not the case, we simply take $\mathcal{D}$ to be the union of all node domains. This need not be done in *practice*, but we use it for simplicity of presentation.

**Definition 1** *A WBFDAG is a 4-tuple $(G, c, r, \mathcal{E})$, where:*

1. $G$ is a connected directed acyclic graph, $G = (V, E)$.
2. $r$ is a function from $V$ to $\mathcal{F}$, called the *label*. If a node $v$ has $k$ immediate predecessors, then the domain of $r(v)$ is $\mathcal{D}'^k$. We use the notation $r_v$ to denote the function $r(v)$
3. $c$ is a function from $(V, \mathcal{D}')$ to the non-negative reals, called the *cost* function.
4. $\mathcal{E}$ is a pair $(s, d)$, the *evidence*. $s$ is a sink node in $G$ and $d$ is the value in $\mathcal{D}'$ *assigned* to $s$.

**Definition 2** *An assignment for a WBFDAG is a function $f$ from $V$ to $\mathcal{D}'$. An assignment is a (possibly partial) model iff the following conditions hold:*

1. If $v$ is a root node then $f(v) \in \mathcal{D}'$.
2. If $v$ is a non-root node, with immediate predecessors $\{v_1, ..., v_k\}$, then $f(v) = r_v(f(v_1), ..., f(v_k))$.

Intuitively, an assignment is a model if the node functional constraints are obeyed everywhere in the WBFDAG. That is, each node can only assume a value dictated by the values of its parents and its label.

**Definition 3** *A model for a WBFDAG is satisfying iff $f(s) = d$.*

**Definition 4** *The cost of an assignment $A$ for a WBFDAG is the sum*

$$C = \sum_{v \in V} c(v, f(v))$$

The Best Selection Problem (BSP) is the problem of finding a minimal cost (not necessarily unique) satisfying model for a given WBFDAG. We examine the BSP in [Charniak and Shimony, 1990]. In that paper, we proved that BSP is NP-hard. We noted there, however, that using standard best first search, we have found minimal cost satisfying (partial) models relatively efficiently.

We now show how to construct a WBFDAG from a Bayesian network, where we make the assumption that only one sink node is an evidence node, and the evidence is of the form "node assumes single value". We then show that the solution to the BSP on the WBFDAG provides us with a MAP assignment for the Bayesian network, and vice versa. Later, the above limitation on evidence nodes is relaxed.

We construct the WBFDAG from the Bayesian network via a local operator on nodes and their immediate predecessors[3]. The domain we use for the WBFDAG is $\mathcal{D}' = \mathcal{D} \cup \{T, F, U\}$. For each root node $u$, construct a node $u'$ (the *image* of $u$) with $|\mathcal{D}|$ parents $u'_i$ (see figure 1), and costs $c(u'_i, T) =$

---

[3]Henceforth, we will use the term "parents" to denote "immediate predecessors".

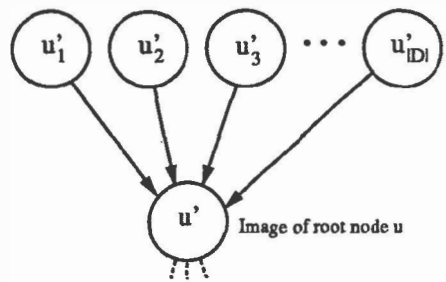

Figure 1: WBFDAG Segment for a Root Node

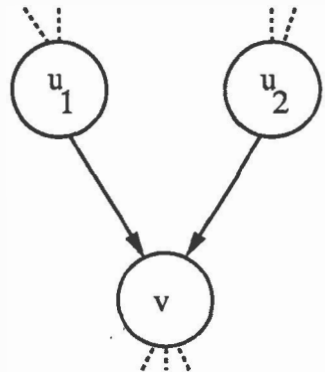

Figure 2: Example Belief Network Segment

$-log(P(f(u) = L_i))$, $c(u'_i, F) = 0$, and $\infty$ cost for all other values[4].

The label $r_{u'}$ of $u'$ is defined as follows:

$$r_{u'} = \begin{cases} L_i & \exists! i.\ f(u'_i) = T \land \\ & \forall j.\ j \neq i \to f(u'_j) = F \\ U & \text{otherwise} \end{cases}$$

Intuitively, $u'$ is an exclusive-or node. For all $d \in \mathcal{D}'$, set $c(u', d) = 0$.

For non-root nodes, the construction is more complicated (consider the belief network segment of figure 2, and the corresponding WBFDAG segment of figure 3 as we describe the construction).

For each non-root node $v$ with in-degree $k$ and parents $\mathcal{U} = \{u_1, ..., u_k\}$ in the Bayesian network, do the following:

1. For each assignment $(d_0, d_1, ..., d_k)$ of values in $\mathcal{D}$ to $\mathcal{U}$ and $v$ (where $d_0 = f(v)$ and $d_i = f(u_i)$), do the following:

(a) Construct a root node $u$ such that $c(u, T)$ is

$$-log(P(f(v) = d_0 | f(u_1) = d_1, ... f(u_k) = d_k))$$

and $c(u, F) = 0$. The cost of all other values for the node is $\infty$.

---

[4]We use the assignment function, $f$, for nodes in the belief net as well as for the WBFDAG. Its meaning in this case should be obvious.

(b) Construct a node $w$ with parents $\mathcal{U}'$ (i.e. the images of the nodes in $\mathcal{U}$) and $u$, and with label function $r_w$, as follows:

$$r_w(d'_1, ... d'_k, d'_u) = \begin{cases} T & d'_u = T \wedge \forall i.\ d'_i = d_i \\ F & \text{otherwise} \end{cases}$$

2. Construct a node $v'$ with $n^{k+1}$ parents, the nodes constructed above. Define $r_{v'}$, the label of $v'$, as follows:

$$r_{v'} = \begin{cases} d(w) & \exists! w.\ w \in parents(v') \wedge \\ & \qquad f(w) = T \\ U & \text{otherwise} \end{cases}$$

where $d(w)$ is the value of $v$ for which $w$ was constructed in step 1b. For all $d \in \mathcal{D}'$, set $c(u', d) = 0$. Intuitively, $r_{v'}$ gets a non-$U$ value just in case exactly one of the parents, $w$ is $T$. We call the node $v'$ constructed in this step the image of $v$ (thus, we call $v$ the *inverse image* of $v'$).

In our example, the belief network segment, with nodes $u_1, u_2, v$, all 2-state nodes, as shown in figure 2, is transformed into the network of figure 3, where the probabilities used to determine the costs of the new root nodes are shown (the actual costs are negative logarithm of the probabilities shown)[5].

The evidence node in the belief network is treated as follows: set $s$ to be the node which is the image of the evidence node, and $d$ to the value of the evidence node.

**Theorem 1** *All minimal cost satisfying models for the WBFDAG induce MAP assignments given the evidence on the Bayesian network.*

Proof outline: we show that any satisfying model for the WBFDAG induces a unique assignment to the nodes of the Bayesian network. We then show that a minimal cost satisfying model for the WBFDAG induces a maximum probability assignment for the Bayesian network.

- The node $s$ can only get a value equal to $d$ if *exactly* one of its parents, $w$, has value $T$, and all others have value $F$. This can happen only if all the parents of each $w$ are assigned values different from $U$. These parents of $w$ are exactly the images of the parents of the inverse image of $s$ (with one new "cost" node constructed in step 1a). Proceeding in this manner to the roots, all image nodes are assigned values in $\mathcal{D}$ in any satisfying model for the WBFDAG. Using exactly these values for the reverse image nodes, we get a unique assignment for the belief network.

- The cost $C$ of a satisfying model is exactly the negative logarithm of the probability of the assignment it induces on the Bayesian network.

To see this, consider the following property of Bayesian networks (see Pearl's book, [Pearl, 1988]). The probability distribution of Bayesian networks can be written as:

$$P(v_1, ..., v_n) = \prod_{v \in V} P(v | parents(v))$$

But in each layer of image nodes, we select exactly one "cost" node to be $T$. The cost of this node is the negative logarithm of the conditional probability of the node state of node $v$ given the state of its parents *in the model*. Now since summing costs is equivalent to multiplying probabilities, the overall cost of the model is the negative logarithm of the overall probability of the induced assignment.

- Finding the MAP is finding the satisfying model $\mathcal{A}$ that maximizes $P(\mathcal{A}|evidence)$. By the definition of conditional probabilities, the latter is:

$$P(\mathcal{A}|evidence) = \frac{P(\mathcal{A})\ P(evidence|\mathcal{A})}{P(evidence)}$$

where $P(evidence)$ is a constant (we are considering a particular evidence instance). Thus, it is sufficient to maximize the numerator.

But $P(evidence|\mathcal{A})$ is exactly $e^{-c}$, where $c$ is the cost of the node selected in the level of the "grandparents" of the evidence node (in figure 3, if $v'$ were the evidence node, we refer to the level of root nodes labeled with $P(F|FT)$... etc.). The latter is true because $P(evidence|\mathcal{A})$ is equal to $P(evidence|\mathcal{A}')$, where $\mathcal{A}'$ is a partial assignment of $\mathcal{A}$, which only assigns values to the parents of the original evidence node (the same values assigned to them by $\mathcal{A}$).

Likewise $P(\mathcal{A})$ is the exponent of the (negative) cumulative cost selected in the rest of the WBFDAG. Since $e^x$ is monotonically increasing in $x$, minimizing the cost of the assignment for the WBFDAG is equivalent to maximizing probability of the assignment to the Bayesian network, Q.E.D.

We now relax the constraint on the evidence, so that the evidence can consist of *any* partial assignment to the nodes of the Bayesian network. Given such a presentation of evidence, we construct an extra node $s$ (in the WBFDAG), with parents exactly the nodes assigned values in the evidence, and assign it the following label function: the node $s$ gets value $T$ just in case its parents are assigned values exactly as in the evidence, and value $F$ otherwise[6]. We now require that $s$ get value $T$ for a satisfying model (the original constraints on the values of the original evidence nodes can be removed). If the evidence is more than just an assignment of one value to each evidence node, we use the method suggested by Pearl before constructing the WBFDAG (see [Pearl, 1988]).

---

[5]As $v$ is a 2-state node, we do not really need *all* the nodes in figure 3, but we show them anyway, so that the generalisation to the m-state node is self evident.

[6]Essentially, $s$ is now an AND node, used for AND'ing all the evidence.

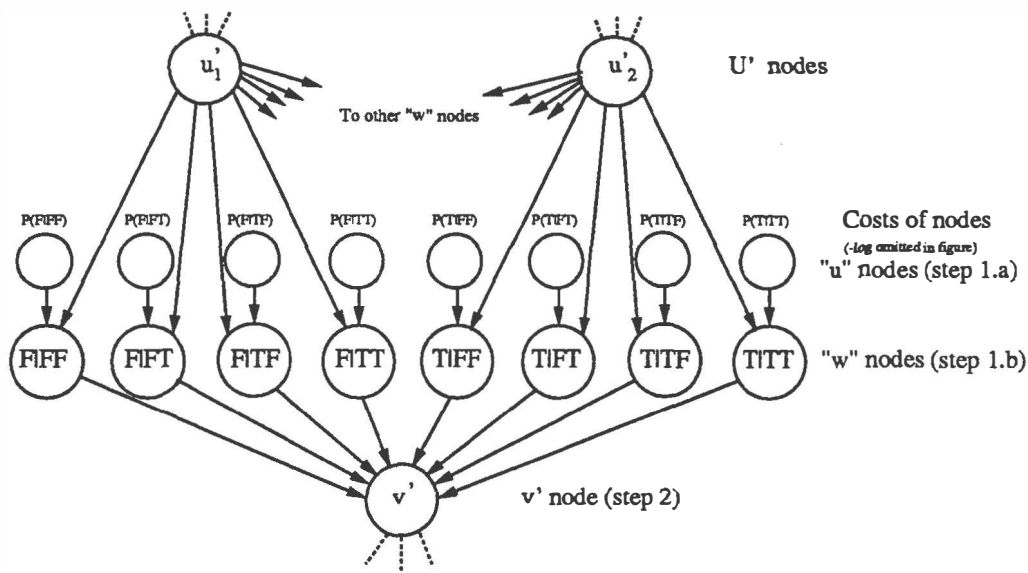

Figure 3: WBFDAG Constructed from Belief Network

## Computing MAPs with WBFDAGs

In the previous section we showed how to construct a WBFDAG from a Bayesian network and evidence such that a minimal cost satisfying model for the WBFDAG induces a maximum probability given the evidence model on the Bayesian network. We now discuss an algorithm for computing MAPs using this construction, and determine its complexity relative to the complexity of the Bayesian network.

Algorithm: compute MAP given evidence $\mathcal{E}$.

1. Construct WBFDAG as in the previous section, where an extra node $e$ is constructed with parents all the nodes in $\mathcal{E}$ if the evidence involves more than just one sink node.

2. Run the best-first search algorithm on the WBFDAG, where the termination condition is a satisfying model[7].

3. Determine the MAP assignment from the model from the roots down.

By letting the best-first search continue after finding the MAP, we can enumerate the assignments to the belief network in order of decreasing probability. We can see that the best-first search algorithm has to run on a graph larger than the Bayesian network, but the size of the WBFDAG is still linear in the size of the Bayesian network. It is, however, exponential in the in-degree of the nodes of the Bayesian network.

If the given Bayesian network has mostly boolean valued nodes, or most of the conditional probabilities are 0 or 1, we can omit most of the construction described above, and save on the size of the WBFDAG. The savings occur because whenever we have a conditional probability of 0, the relevant $u$ and $w$ nodes can be omitted (as the $u$ node has cost $\infty$). Whenever we have a conditional probability of 1, we can essentially omit the $u$ node (and modify the $w$ node label accordingly). In the extreme case, where all non-root nodes in the belief network have only boolean conditional probabilities (i.e. only values of 0 or 1 in their conditional distribution arrays), no special construction is necessary, as shown in [Charniak and Shimony, 1990].

---

[7]At this point we will apply standard best-first-search on AND-OR trees to our WBFDAG to find the minimum cost. Since our WBFDAG is, however, an AND-XOR DAG, not an AND-OR tree, it is, perhaps worth describing the best-first search technique to show why it still applies. Best-first search on AND-OR trees works by starting at the sink and constructing alternative partial solutions. Whenever an OR node with $k$ parents is encountered, we split our partial solution into $k$, each one of which will contain the previous partial solution but now extended to include on of the OR possibilities. Whenever an AND node is encountered, all of its predecessors are added as things we must now handle. If we have a DAG then we must simply check whenever a new node is added to the partial solution that it has not been added before. If it has, it is simply not added the second time. As for the XOR nodes, in fact, best-first-search is commonly used in exclusive-or situations (e.g., graph coloring, where the choice of color for a region is exclusive.) Using the technique in the XOR case is simply a matter of making sure that a variable (region, or random variable) gets only one value (color, or value of the random variable). In our case this is complicated by the seeming possibility that we assign random variables $v_1 = T$ and $v_2 = F$, whereas in our distribution we have it that $v_1 \Rightarrow v_2$. In fact, this cannot occur, but we omit the proof.

The best-first search will run in linear time on poly trees, assuming that the correct bookkeeping operations are made (i.e. the best assignment cost for the ancestors of a node is kept at every node, for every possible value assigned to the node). This is true because once we have these least-cost values for a node, there is no need to expand its ancestors again. In fact, Pearl's algorithm for computing MAPs relies on this property (see [Pearl, 1988]). Thus, if the poly tree belief network has only boolean distribution for all nodes, then, because the WBFDAG constructed is also a poly tree, we have an algorithm that runs in time linear in the size of the network[8]. When finding next-best MAPs, however, we can no longer rely on the above property, and thus can no longer guarantee linear time.

Unfortunately, for general poly tree belief networks, once we construct the WBFDAG, we no longer have a poly tree! We can show, however, that we still have an algorithm with running time linear in the size of the network. Note that the WBFDAG is still separable into components, where the separating nodes are the images of the nodes of the original poly tree. Also, from the "cost" nodes constructed for a certain node $v$, only *one* is selected to be assigned $T$. Using these constraints, the algorithm still runs in time linear in the size of the belief network.

Finally, our algorithm can be easily modified to compute certain partial MAPs. If we are only interested in assignments to some subset of root nodes in the belief network (the root nodes could represent diseases in medical diagnosis, for instance), all we need to do is set to 0 the costs of all root nodes in the WBFDAG that are *not* parents of images of root nodes.

### Implementation

The algorithm has been implemented for the belief networks generated by WIMP (see [Charniak and Goldman, 1988]), where most nodes have only two states and many conditional probabilities are either 0 or 1. The results are rather optimistic, as partial MAPs were computed faster than evaluating posterior probabilities for the nodes of the same network given the same evidence. For that experiment, a very trivial admissible heuristic was used[9], and it is certainly reasonable to hope that a better admissible heuristic will improve performance even further. No conclusive timing tests have been conducted, however.

In WIMP, only one set of evidence is used per belief network, as networks are constructed on the fly as new evidence comes in. If we need to use the same belief net with different evidence, however, the WBFDAG can be used again (with minor changes to cater for the different evidence). It is possible that many of the best-first search computations are also re-usable, but we did not try to do that, because it was not useful for our domain.

We have an improved implementation of the algorithm, where the assumption that 0 and 1 conditional probabilities abound is dropped. The implementation avoids the actual construction of the extra nodes, even though conceptually the nodes are still there. This version of the algorithm exploits cases where many adjacent entries in the conditional distribution array are equal, but not necessarily 0 or 1. Using this property, many of the (virtual) $w$ nodes are collapsed together, and likewise the $u$ nodes. Advantages of this method over the method described earlier in this section is that it facilitates treatment of noisy ORs and ANDs (and many other types of nodes), as well as pure ORs and ANDs. Detailed discussion of the modified algorithm is outside the scope of this paper, but see [Shimony, 1990].

### Conclusions and Future Research

We demonstrated a new algorithm for finding MAPs for belief networks, using a best-first search on a WBFDAG constructed from the given belief network and the evidence. We also made it evident that any algorithm that solves the BSP on the WBFDAG is also good for finding MAPs, and thus any possible improvement there will also make finding the MAP more efficient. We showed an effective way of finding successive next-best assignments for Bayesian networks.

Further research is needed for improving the best-first search for finding the minimal cost model for the WBFDAG, which may be possible for limited networks generated for special cases, such as in the case of networks generated by WIMP. Also, more empirical data comparing this algorithm to existing ones, is necessary. For that purpose, we intend to test the algorithm on randomly generated belief networks.

---

[8] Pearl's algorithm for finding MAP is also efficient (time linear in the size of the network), for poly trees. In some cases (i.e. if local best assignments also happen to be global best assignments) our algorithm will avoid many operations that Pearl's algorithm has to perform, but in general the running times will be equal.

[9] Whereby the cost of the complete assignment is evaluated at the cost collected until now.

# Session 4:

## First Poster Session

Reducing Uncertainty in Navigation and Exploration
*K. Bayse, M. Lejter, and K. Kanazawa*

Ergo: A Graphical Environment for Constructing Bayesian Belief Networks
*I. Beinlich and E. Herskovits*

Decision Making with Interval Influence Diagrams
*J.S. Breese and K.W. Fertig*

A Randomized Approximation Algorithm of Logic Sampling
*R.M. Chavez and G.F. Cooper*

Occupancy Grids: A Stochastic Spatial Representation for Active Robot Perception
*A. Elfes*

Time, Chance, and Action
*P. Haddawy*

A Dynamic Approach to Probabilistic Inference Using Bayesian Networks
*M.C. Horsch and D. Poole*

Approximations in Bayesian Belief Universe for Knowledge Based Systems
*F. Jensen and S.K. Andersen*

Robust Inference Policies: Preliminary Report
*P.E. Lehner*